\documentclass[sigconf]{acmart}
\usepackage{subfigure}
\usepackage{algorithm}
\usepackage{algorithmic}
\usepackage{amsthm}
\AtBeginDocument{%
  \providecommand\BibTeX{{%
    \normalfont B\kern-0.5em{\scshape i\kern-0.25em b}\kern-0.8em\TeX}}}

\setcopyright{acmcopyright}
\copyrightyear{2018}
\acmYear{2018}
\acmDOI{10.1145/1122445.1122456}

\acmConference[Woodstock '18]{Woodstock '18: ACM Symposium on Neural
  Gaze Detection}{June 03--05, 2018}{Woodstock, NY}
\acmBooktitle{Woodstock '18: ACM Symposium on Neural Gaze Detection,
  June 03--05, 2018, Woodstock, NY}
\acmPrice{15.00}
\acmISBN{978-1-4503-XXXX-X/18/06}

\acmSubmissionID{440}

\begin{document}

\title{Imitate The World: A Search Engine Simulation Platform}

\author{Yongqing Gao}\authornote{Both authors contributed equally to the paper.}
\email{gaoyongqing@smail.nju.edu.cn}
\affiliation{%
  \institution{Nanjing University}
  \country{China}
}
\author{Guangda Huzhang}\authornotemark[1]
\email{guangda.hzgd@alibaba-inc.com}
\affiliation{%
  \institution{Alibaba Group}
  \country{China}
}
\author{Weijie Shen}
\email{shenweijie93@smail.nju.edu.cn}
\affiliation{%
  \institution{Nanjing University}
  \country{China}
}
\author{Yawen Liu}
\email{liuyw@smail.nju.edu.cn}
\affiliation{%
  \institution{Nanjing University}
  \country{China}
}
\author{Wen-Ji Zhou}
\email{eric.zwj@alibaba-inc.com}
\affiliation{%
  \institution{Alibaba Group}
  \country{China}
}
\author{Qing Da}
\email{daqing.dq@alibaba-inc.com}
\affiliation{%
  \institution{Alibaba Group}
  \country{China}
}
\author{Yang Yu}
\email{yuy@nju.edu.cn}
\affiliation{%
  \institution{Nanjing University}
  \country{China}
}

% \affiliation{%

% \affiliation{%

% \affiliation{%

% \affiliation{%

% \affiliation{%

% \affiliation{%

% \affiliation{%

\renewcommand{\shortauthors}{Trovato and Tobin, et al.}

\begin{abstract}
Recent E-commerce applications benefit from the growth of deep learning techniques. However, we notice that many works attempt to maximize business objectives by closely matching offline labels which follow the supervised learning paradigm. This results in models obtain high offline performance in terms of Area Under Curve (AUC) and Normalized Discounted Cumulative Gain (NDCG), but cannot consistently increase the revenue metrics such as purchases amount of users.  Towards the issues, we build a simulated search engine AESim that can properly give feedback by a well-trained discriminator for generated pages, as a dynamic dataset. Different from previous simulation platforms which lose connection with the real world, ours depends on the real data in AliExpress Search: we use adversarial learning to generate virtual users and use Generative Adversarial Imitation Learning (GAIL) to capture behavior patterns of users. Our experiments also show AESim can better reflect the online performance of ranking models than classic ranking metrics, implying AESim can play a surrogate of AliExpress Search and evaluate models without going online.

\end{abstract}




\maketitle

\section{Introduction}
With increasing research activities in deep learning theory and practices, Learning-to-Rank (LTR) solutions rapidly evolve in many real-world applications. As the main component of an online system in E-commerce, LTR models strongly connect to the business profit. However, industrial LTR studies meet two stubborn issues. First, as the recent RecSys work~\cite{dacrema2019we} reveals, many works cannot be perfectly reproduced. Second, even we successfully reproduce the performances of proposed methods in a specified task, it is hard to promise the same performance in another task, as well as in the online environments. Therefore, it is desired for researchers to have a public platform for E-commerce LTR  evaluation.

\begin{figure}[h]
  \includegraphics[scale=0.283]{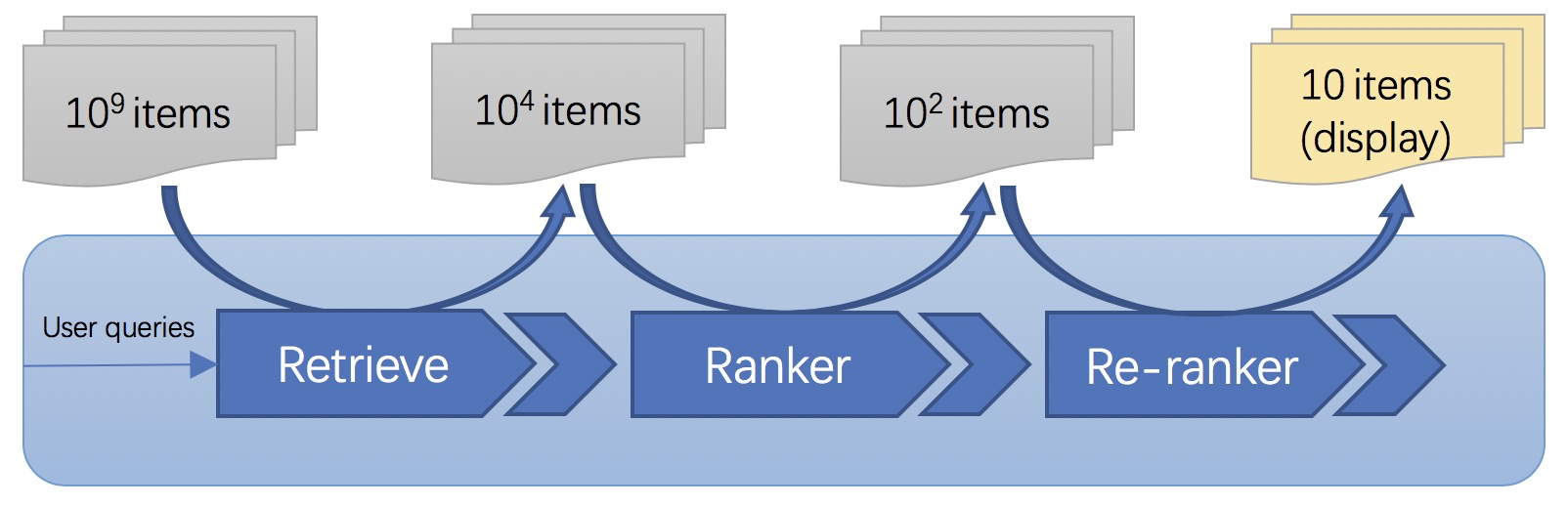}
  \caption{A typical process of an industrial search engine.}
  \label{fig:teaser}
\end{figure}

\begin{figure*}[ht]
  \includegraphics[width=0.99\linewidth]{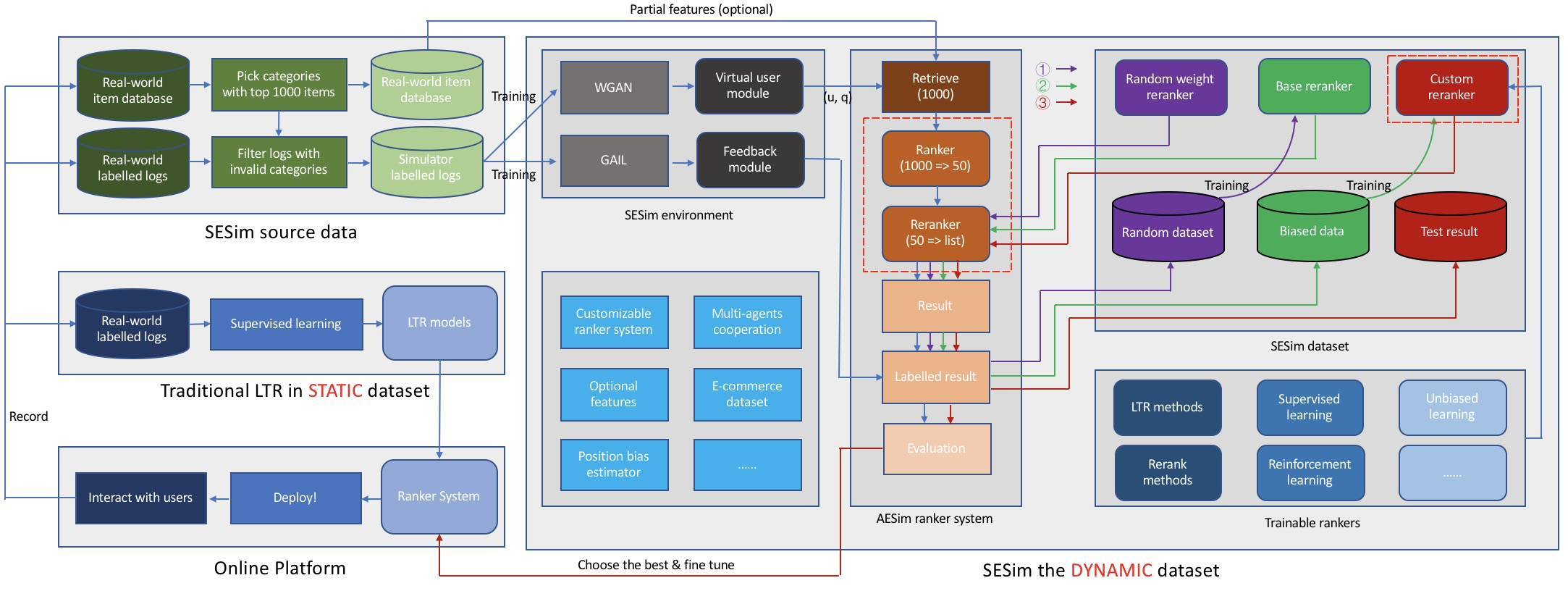}
  \caption{The workflow of AESim.}
  \label{fig:teaser}
\end{figure*}

A typical process of industrial search engines contains three stages to produce a display list from a user query. A search engine first retrieves related items with intend of the user (i.e. the user query), then the \emph{ranker} ranks these items by a fine-tuned deep LTR model, finally, the \emph{re-ranker} rearranges the order of items to achieve some businesses goals such as diversity and advertising. Our proposed simulation platform AESim contains these three stages. We replace queries with category indices in our work, therefore AESim can retrieve items from a desensitized items database by the category index. After that, a customizable ranker and a customizable re-ranker produce the final item list. AESim allows us to study joint learning of multiple models, we left it as future work and focus on the correct evaluation for a single model.

Besides the set of real items, two important modules make AESim vividly reflect the behaviors of real users. \emph{Virtual user module} aims at generating embeddings of virtual users and their query, and it follows the paradigm of Wasserstein Generative Adversarial Network with  Gradient Penalty (WGAN-GP). \emph{Feedback module} inputs the display list and the information of the user, then outputs the feedback of users on the display list. To model the decision process of users, we train the feedback module by Generative Adversarial Imitation Learning (GAIL). For diversifying behaviors, we consider clicking and purchasing, which are two of the most important feedback of users in E-commerce.

The contribution of AESim includes:
\begin{itemize}
\item As far as we know, AESim is the first E-commerce simulation platform generated by imitating real-world users. 
\item AESim can be used as a fair playground for future studies on E-commerce LTR researches.
\item Our online A/B testings show that AESim can reflect online performance without online interaction.
\end{itemize}

\section{Related Works}
Generally, most Learning-to-rank (LTR) models are partitioned into three groups: point-wise models, pair-wise models, and list-wise models. These methods have different forms of loss functions. Point-wise models~\cite{cossock08point,li07point,hu2008collaborative} focus on an individual classification or regression task. The loss of pair-wise models~\cite{joachims02pair,Burges:ranknet,burges2010ranknet,rendle2012bpr,mao2020www} include pairs of scored items, and it is computed by the relative relationship of their scores and labels. List-wise models~\cite{cao07list,xia08list,ai2018learning,yu2020collaborative} score items to optimize the holistic metrics of lists. Practically, all these models give item scores and the online system will rank items straightforwardly by the scores. However, evaluating models by historical data is problematic, which may lead the online-offline inconsistency~\cite{huzhang2020aliexpress,rohde2018recogym,beel2013comparative,rossetti2016contrasting,mcnee2006being}.

To correctly evaluate a model without going online, a simulation platform is necessary to give a dynamic response for a newly generated list. There are several simulation platform for search engines and recommender systems, such as Virtual-Taobao~\cite{shi2019virtual}, RecSim~\cite{ie2019recsim} and RecoGym~\cite{rohde2018recogym}. However, Virtual-Taobao cannot give an evaluation for a complete list. RecSim and RecoGym can evaluate reinforcement learning models, but they lose the connection to real-world application.   Our model follows generative adversarial imitation learning (GAIL)~\cite{goodfellow15gan} , which has been examined to be a better choice for imitation learning ~\cite{ho16gail,finn2016guided,shi2019virtual}, to learn the patterns of real users.

\section{The Proposed Framework}
AESim includes an item database, a virtual user module, a feedback module, a customizable ranker system, and generated datasets. It can test LTR algorithms with a straightforward evaluation and can test de-biasing methods in a pure offline environment. The item database contains millions of selected active items and these items are categorized with their category indices. To train and evaluate a ranker model, AESim first prepares the training set and the testing set of labeled lists by the virtual user module (generate queries), a complete ranker system (generate final lists), the feedback module (generate feedback of virtual users).  With the training set, we can train new ranker models and produce results for the testing set. Finally, we use the feedback module again to examine the true performance of the ranker model.

\begin{figure*}[t]
    \subfigure[Overview of discrete features.]{
		\label{level.sub.1}
		\includegraphics[width=0.36\linewidth]{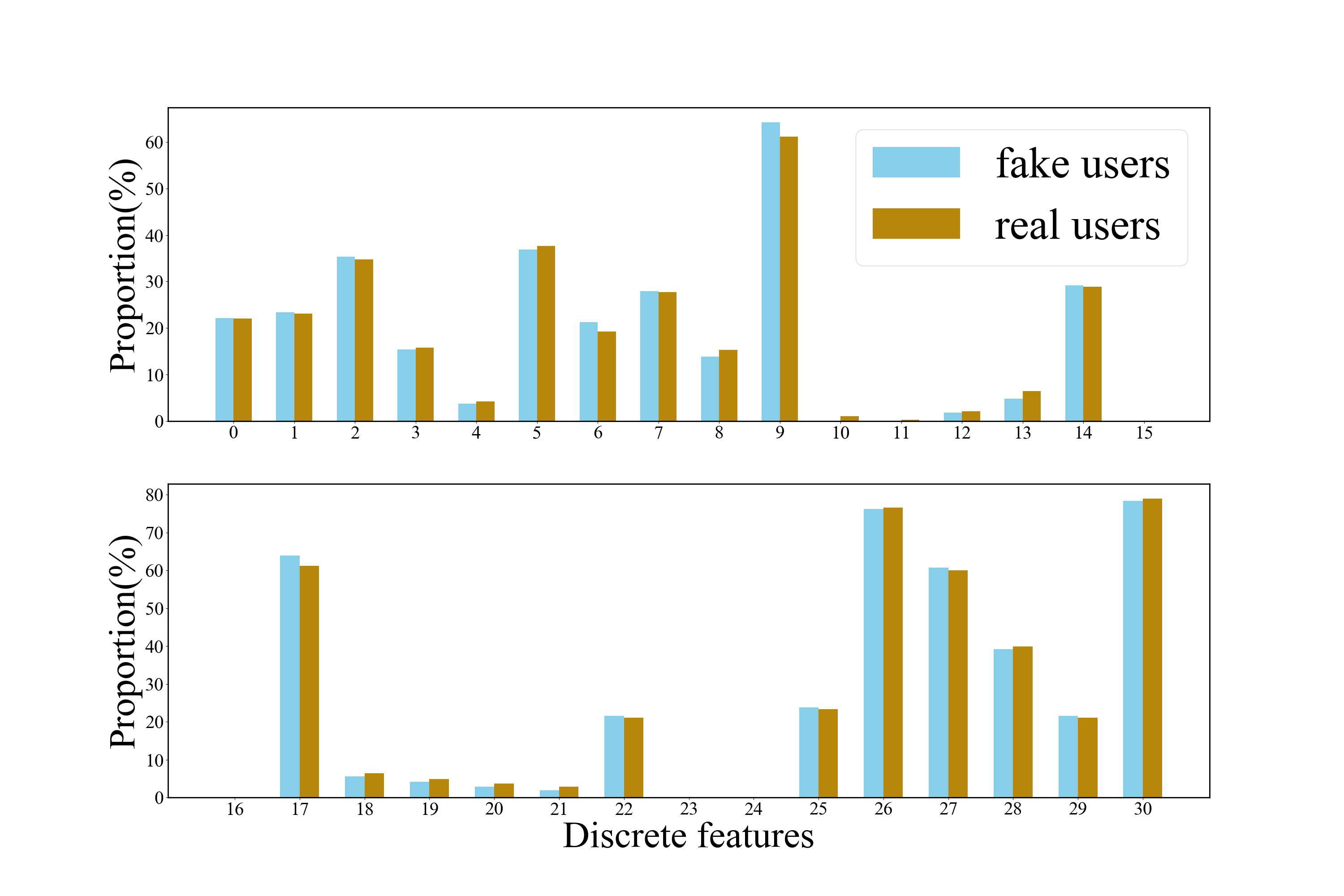}}
    \subfigure[Overview of dense features.]{
		\label{level.sub.2}
		\includegraphics[width=0.36\linewidth]{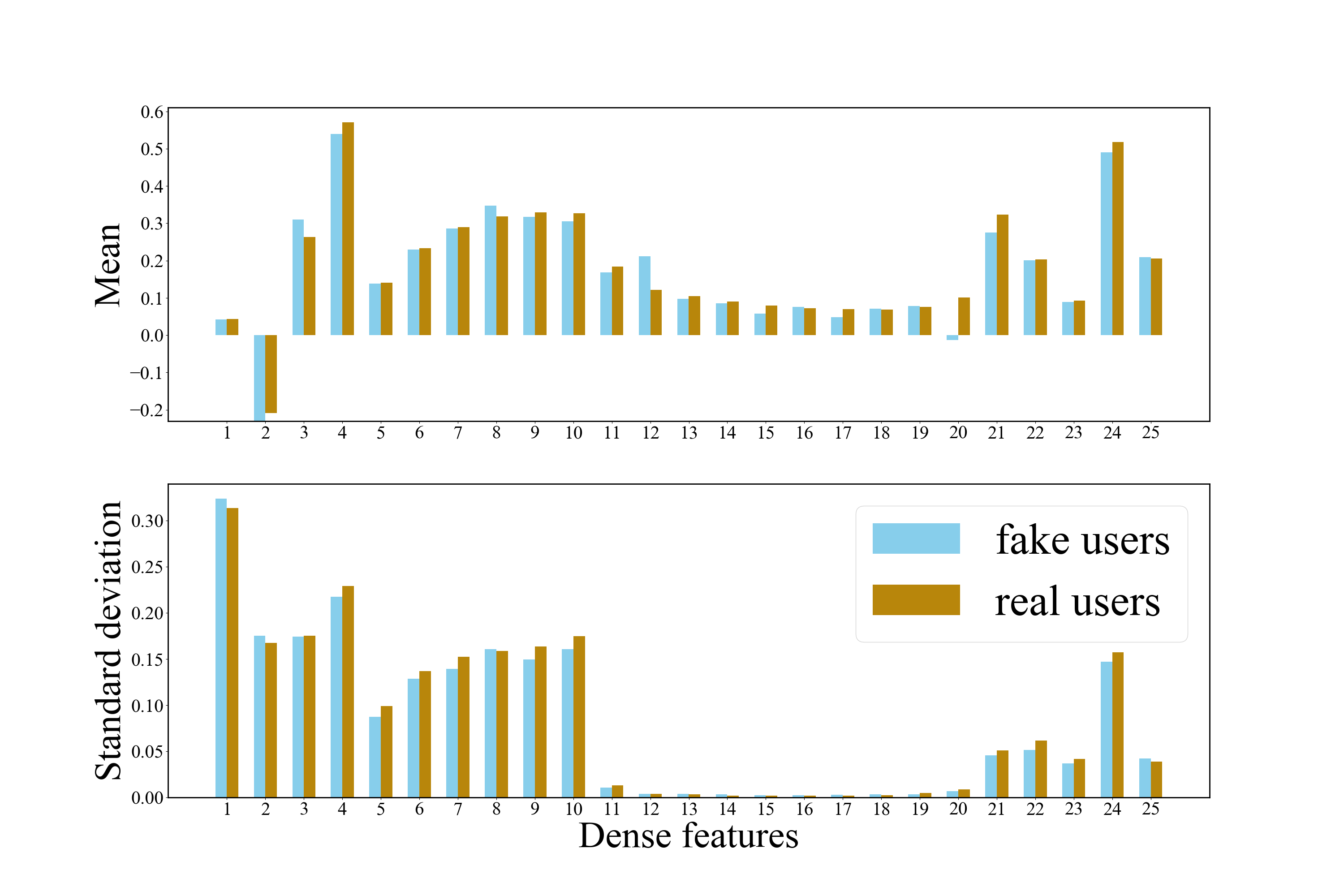}}
    \subfigure[Distribution of top 10 queries.]{
		\label{level.sub.3}
		\includegraphics[width=0.25\linewidth]{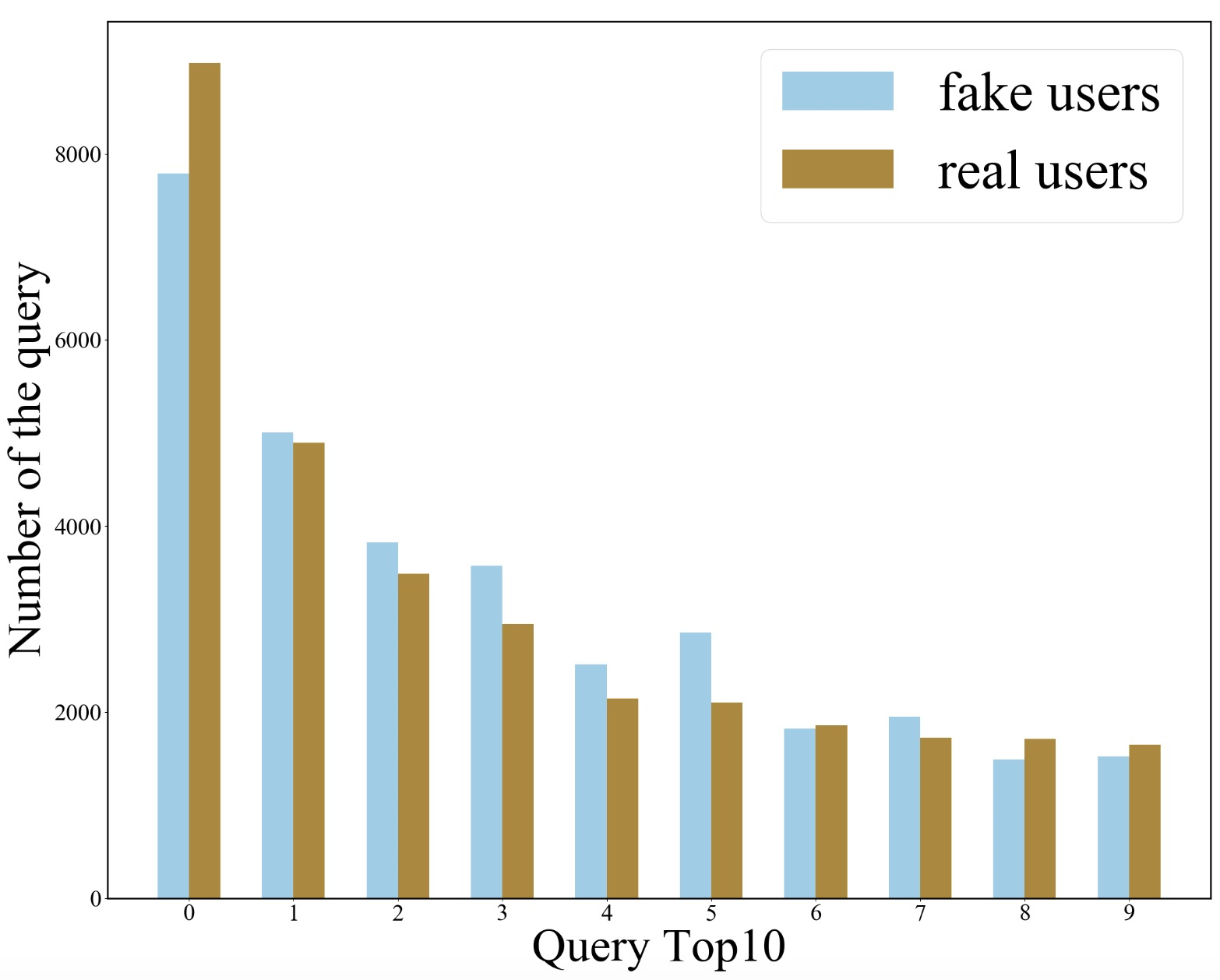}}
  \caption{The simulation effect of AESim.}
  \label{fig:sim}
\end{figure*}

\subsection{Virtual User Module}
Virtual User Module contains a generator and a discriminator which are trained following WGAN-GP. The generator aims at generating features of users and his query which are similar to the real records. The discriminator tries to distinguish the fake(generated) and real pairs of users and queries and guides the generator to reach its objective. The loss function of the discriminator is 
\begin{equation}
\begin{aligned}
    L({\theta_D}) &= (\mathbb{E}_{\widetilde{x}\sim\mathbb{P}_f^{\theta_G}}[D(\widetilde{x}|\theta_D)] - \mathbb{E}_{x\sim\mathbb{P}_r}[D(x|\theta_D)]) \\&+ \lambda \mathbb{E}_{\hat{x}\sim\mathbb{P}_f^{\theta_G}\cup \mathbb{P}_r}[(\left\| D(\hat{x}|\theta_D)\right\|_2-1)^2]
\end{aligned}
\label{eq:wgangp}
\end{equation}

The generator tries to minimize the following loss:
\begin{equation}
\begin{aligned}
    L({\theta_G}) = -\mathbb{E}_{\widetilde{x}\sim\mathbb{P}_f^{\theta_G}}[log(D(\widetilde{x}|\theta_D))]
\end{aligned}
\end{equation}

\begin{figure}[h]
  \includegraphics[scale=0.21]{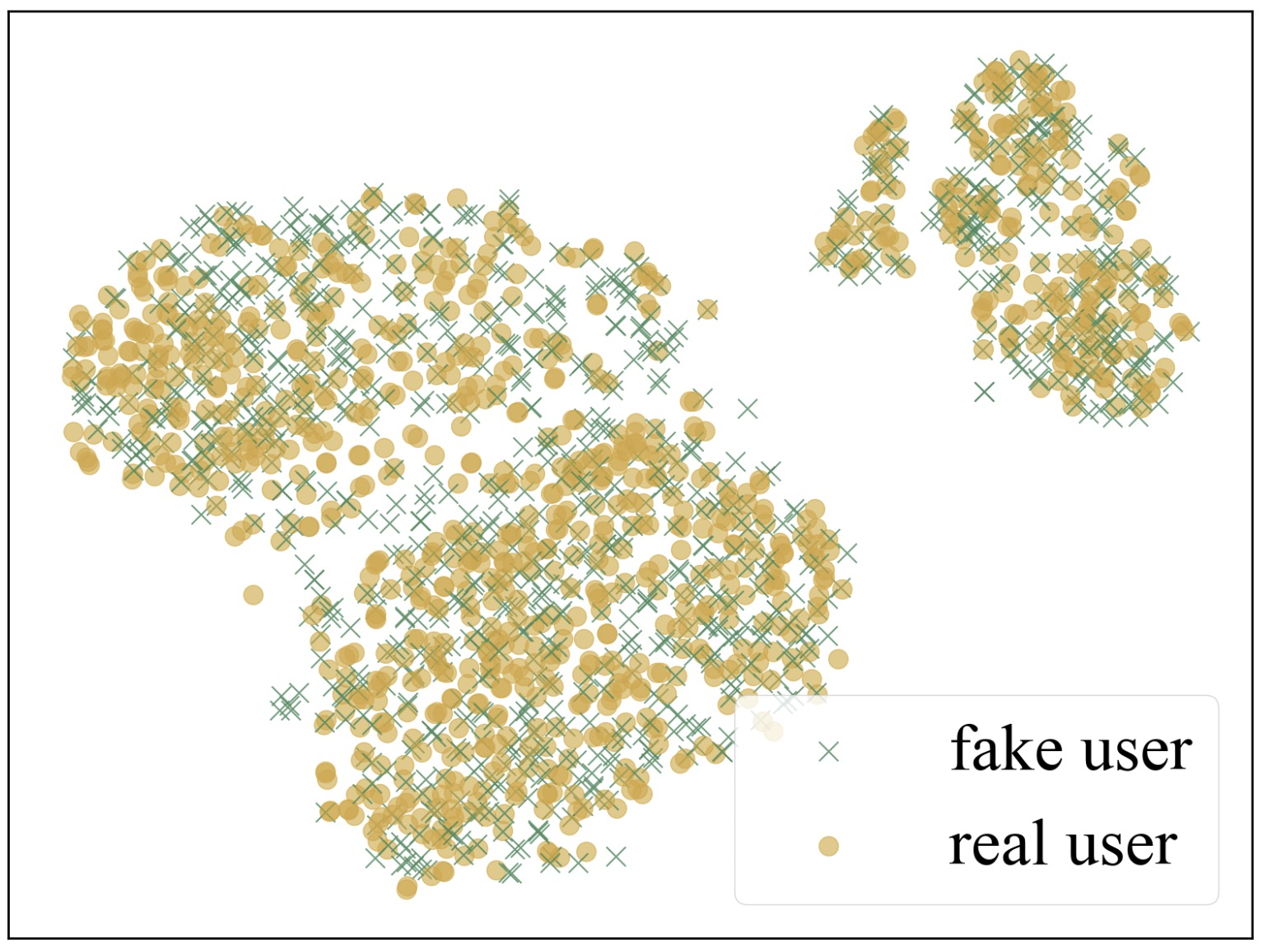}
  \caption{A typical process of an industrial search engine.}
  \label{fig:tsne}
\end{figure}

Sample $\widetilde{x}$ is generated by the generator and $\mathbb{P}_f^{\theta_G}$ is the distribution of outputs of the generator with parameters $\theta_G$, and sample $x$ is the real sample from the real sample distribution $\mathbb{P}_r$. The third term in Equation~\ref{eq:wgangp} is the core trick of WGAN-GP.
In our design, the structure of the evaluator and the generator are multi-layer perceptions with hidden layer sizes $[128, 64, 32]$. To visualize the similarity between real data and generated data, we plot the distribution of features of users in Figure~\ref{fig:sim}. On the other hand, we consider the joint distribution of users and queries and use TSNE to plot them in a plane. Figure~\ref{fig:tsne} shows the generated data have similar patterns to the real data. Both figures imply our generated virtual users can hardly be distinguished from real users.

\subsection{Ranker System of AESim}
The process of ranking in AESim is similar to real search engines. After the virtual user module generates a user-query pair, the ranker system inside AESim starts to compute the final display list. First, it retrieves $1000$ items from the item dataset with the query, which is translated into the category index in our work. Then, the ranker (a point-wise model) scores the items and sends the top $50$ of them to the re-ranker, and the re-ranker decides the final order of items. Finally, AESim evaluates the output of the ranker system by the feedback module.

\subsection{Feedback Module}
Our feedback module has a classic sequence-to-sequence structure and rewards each item by imitating the behaviors of real users. We train its parameters follows GAIL: a discriminator in included to distinguish how the generated behaviors are close to the behaviors of real users. In our work, we also try to use WGAN-GP to generate feedback. The outputs of the feedback module following GAIL are much more similar to real behaviors than the one following WGAN-GP. In GAIL, the gradient of discriminator has the following form:
\begin{equation}
\label{eq:dupdate}
\hat{\mathbb{E}}_{\tau_f^{\theta_G}}\left[\nabla_{\theta_D} \log \left(D(s, a|\theta_D)\right)\right]+\hat{\mathbb{E}}_{\tau_r}\left[\nabla_{\theta_D} \log \left(1-D(s, a|\theta_D)\right)\right]
\end{equation}
Here $\tau_f^{\theta_G}$ is the generated trajectory with parameters $\theta_G$ and $\tau_r$ is the real trajectory. State $s$ and action $a$ are included in the trajectories. The parameters of generator $\theta_G$ is updated with reward function $log(D(s, a|\theta_D))$ using the TRPO rule.
% \begin{equation}
% \begin{aligned}
% \label{eq:gail}
% \hat{\mathbb{E}}_{\tau_f}\left[\nabla_{\theta_D} \log \left(D(s, a|\theta_D)\right)\right]+\hat{\mathbb{E}}_{\tau_r}\left[\nabla_{\theta_D} \log \left(1-D(s, a|\theta_D)\right)\right]
% \end{aligned}
% \end{equation}

%The main difference between them is that WGAN-GP focuses on similarity of (state, action) pairs but GAIL considers similarity of the complete trajectory. 

\begin{table*}[ht]
  \small
\resizebox{.96\textwidth}{!}{
  \begin{tabular}{l|llll|llll}
  \toprule 
  \multicolumn{1}{c|}{}    & \multicolumn{4}{c|}{\textbf{No de-biasing}}            &  \multicolumn{4}{c}{\textbf{De-biasing}}\\ 
  \multicolumn{1}{c|}{\textbf{Method}}    &  \multicolumn{1}{c}{\textbf{GAUC}}            & \multicolumn{1}{c}{\textbf{NDCG}} & \multicolumn{1}{c}{\textbf{MAP}}& \multicolumn{1}{c|}{\textbf{AESim}} & \multicolumn{1}{c}{\textbf{GAUC}}            & \multicolumn{1}{c}{\textbf{NDCG}} & \multicolumn{1}{c}{\textbf{MAP}}& \multicolumn{1}{c}{\textbf{AESim}} \\ \hline
  %  Dictator &   &     &      &     &        &      &  &     \\ 
  %  Copeland &   &     &      &     &        &      &  &      \\ \hline
Point-wise & 0.806283 & 0.623264 & 0.025052 & 0.003089 & 0.805345 & 0.620394 & 0.024876 & 0.003074 \\
% PairWise-hinge & 0.807165 & 0.623744 & 0.025091 & 0.003092 & - & - & - & - \\
Pair-wise & 0.805478 & 0.621492 & 0.024959 & 0.003080 & 0.804045 & 0.618290 & 0.024746 & 0.003084 \\
ListMLE & 0.799506 & 0.626811 & 0.025345 & 0.002984 & 0.794266 & 0.629366 & 0.025500 & 0.002952 \\
Group-wise & 0.806052 & 0.622164 & 0.025001 & 0.003052 & 0.805424 & 0.619805 & 0.024861 & 0.003075 \\
DLCM & 0.807749 & 0.634064 & 0.025770 & 0.002657 & 0.807156 & 0.633718 & 0.025757 & 0.002615 \\
  \bottomrule
  \end{tabular}}
\vspace{0.8em}
\caption{The model results in AESim.}
\label{tab:AESimresult}
 \end{table*}

\begin{figure}[h]
  \includegraphics[scale=0.135]{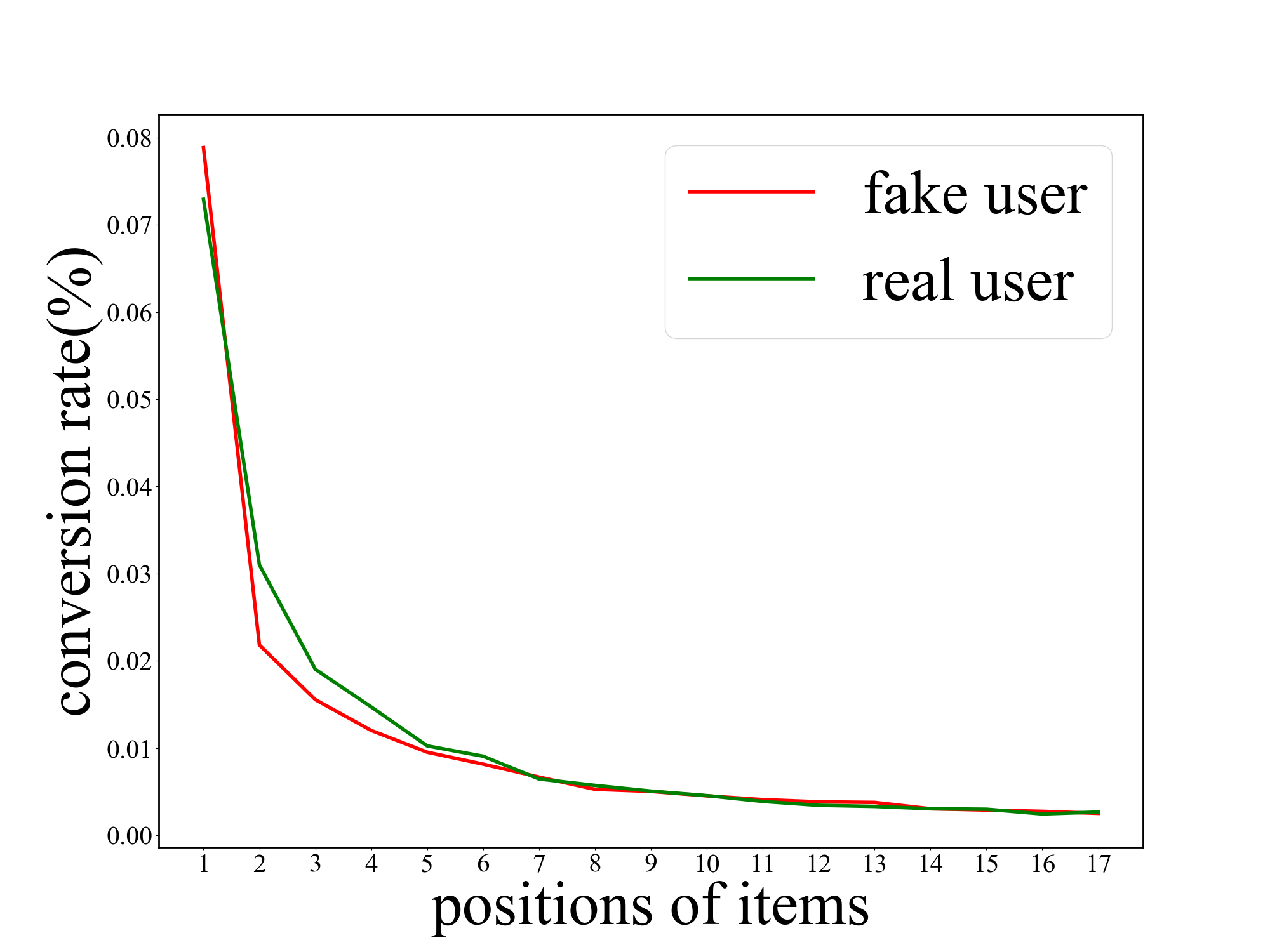}
  \caption{The purchase trend of users at each position in real and fake scenarios. }
  \label{fig:trend}
\end{figure}

\begin{table*}[ht]
\centering
\resizebox{.99\textwidth}{!}{
\vspace{0.8em}
\begin{tabular}{c|c|cccccccccc}
\toprule &\textbf{AESim}  & \textbf{Day 1} & \textbf{Day 2} & \textbf{Day 3} & \textbf{Day 4} & \textbf{Day 5} & \textbf{Day 6} & \textbf{Day 7} & \textbf{Day 8} & \textbf{Day 9} & \textbf{Day 10} \\
\hline
Point-wise & +0.00\% & +0.54\% & +0.59\% & +1.11\% & +0.70\% & +0.62\% & +1.15\% & +0.29\% & -0.24\% & +0.12\% & +0.31\%\\
Pair-wise &  -0.26\% & -0.39\% & -0.05\% & +1.02\% & +0.25\% & +0.39\% & +1.34\% & +0.86\% & -0.60\% & -0.19\% & -0.85\%\\
ListMLE & -1.00\% &  -2.15\%& -1.51\% & -0.29\% & -1.46\% & -2.57\%&  -2.17\% & -2.24\% & -2.83\% & -2.32\% & -3.81\%\\
DLCM & -17.1\% & -1.34\%&-0.42\%&+0.19\%&-1.52\%&-2.72\%&-1.39\%&-1.95\%&-0.31\%&-0.48\%&-1.36\%\\

\bottomrule
\end{tabular}
}
\vspace{0.8em}
\caption{Online performance and AESim evaluations of models.}
\label{tab:one}
\end{table*}
Compared to previous simulation platforms, ours has a significantly similar purchase trend to real-world applications. Figure~\ref{fig:trend} shows the conversion rate of the purchase at each position with the real feedback and the generated feedback. This property motivates ranker systems to put better items at the top.

\subsection{Dataset Preparation}
To build the dataset for model training and testing, we further need rankers which help the ranker system produce the initial dataset. We first use a \emph{random weight ranker} to generate a random training set for training a \emph{base ranker}. Then we use a base ranker to generate the final training dataset and testing dataset for model evaluation. An important benefit of the above steps is that we can reproduce the sample selection bias issue of offline data.

The training set in AESim is the same as the traditional static dataset for supervised learning models. The main difference appears in the testing phase: AESim can give accurate response for any newly generated list, but static dataset can only use the old feedback from old lists, where the order of items is already changed.

\section{Experiment}
\paragraph{Offline Testing.} We test a point-wise method with a cross-entropy loss, a pair-wise method with a logistic loss~\cite{Burges:ranknet}, the listMLE~\cite{ai2018learning}, and the group-wise scoring  framework~\cite{ai2019learning} (GSF) in AESim, where all these methods use the same MLP (note that GSF contains several isomorphic MLP). We further add DLCM which is expected to have a high offline performance for its complicated structure. 

To include the de-biasing methods, we proceed with a simulation in AESim to swap the first item and the $k$-th items, then observe the change of conversion rate to determine the value of position bias~\cite{joachims2017unbiased}. After that, some of the above methods can add an inverse propensity score to remove the influence brought by position bias.
%list-wise networks and group-wise scoring frameworks to compare their performance on AESim. 
It can be observed in Table~\ref{tab:AESimresult} that GAUC, NDCG and MAP have similar preferences for models but AESim scores give different orders. Especially, DLCM gets the highest GAUC but obtains low AESim scores, which implies that a model with high GAUC may fail to optimize the online performance. 

\paragraph{Online Testing.} To examine that AESim correctly evaluates the models, we put the point-wise model, the pair-wise model, and the ListMLE in our online system. Each model needs to serve a non-overlapping random portion of search queries as a re-ranker. Roughly, each model serves millions of users and produces millions of lists per day. Due to the daily dramatic change of online environments, the difference gap may perform unstably so that we need to consider the overall performance of models. The ten days result in Table~\ref{tab:one} shows the consistency with our offline evaluation for the point-wise model, the pair-wise model, and the ListMLE. However, DLCM is evaluated extremely poor in AESim and its performance is not that bad when serves online. Therefore, we suggest considering AESim as a rough judgment for a model, which may have a gap with the actual performance.

\section{conclusion}
We propose an E-Commerce search engine simulation platform for model examinations, which was a missing piece to connect evaluations of LTR researches and business objectives of real-world applications. AESim can examine models in the simulation E-commerce environment with dynamic responses, and its framework can be easily extended to other scenarios that items and users have different features. We hope to see the development of a dynamic dataset that facilitates industrial LTR researches in the future.

% \section{Acknowledgments}
% This work was supported by Alibaba Group through Alibaba Innovative Research Program.
\bibliographystyle{ACM-Reference-Format}
\bibliography{acmart}


\begin{thebibliography}{25}

%%% the Web address.

\ifx \showCODEN    \undefined \def \showCODEN     #1{\unskip}     \fi
\ifx \showDOI      \undefined \def \showDOI       #1{#1}\fi
\ifx \showISBNx    \undefined \def \showISBNx     #1{\unskip}     \fi
\ifx \showISBNxiii \undefined \def \showISBNxiii  #1{\unskip}     \fi
\ifx \showISSN     \undefined \def \showISSN      #1{\unskip}     \fi
\ifx \showLCCN     \undefined \def \showLCCN      #1{\unskip}     \fi
\ifx \shownote     \undefined \def \shownote      #1{#1}          \fi
\ifx \showarticletitle \undefined \def \showarticletitle #1{#1}   \fi
\ifx \showURL      \undefined \def \showURL       {\relax}        \fi
% invisible to TeX
\providecommand\bibfield[2]{#2}
\providecommand\bibinfo[2]{#2}
\providecommand\natexlab[1]{#1}
\providecommand\showeprint[2][]{arXiv:#2}

\bibitem[\protect\citeauthoryear{Ai, Bi, Guo, and Croft}{Ai
  et~al\mbox{.}}{2018}]%
        {ai2018learning}
\bibfield{author}{\bibinfo{person}{Qingyao Ai}, \bibinfo{person}{Keping Bi},
  \bibinfo{person}{Jiafeng Guo}, {and} \bibinfo{person}{W~Bruce Croft}.}
  \bibinfo{year}{2018}\natexlab{}.
\newblock \showarticletitle{Learning a deep listwise context model for ranking
  refinement}. In \bibinfo{booktitle}{\emph{The 41st International ACM SIGIR
  Conference on Research \& Development in Information Retrieval}}. ACM,
  \bibinfo{pages}{135--144}.
\newblock


\bibitem[\protect\citeauthoryear{Ai, Wang, Bruch, Golbandi, Bendersky, and
  Najork}{Ai et~al\mbox{.}}{2019}]%
        {ai2019learning}
\bibfield{author}{\bibinfo{person}{Qingyao Ai}, \bibinfo{person}{Xuanhui Wang},
  \bibinfo{person}{Sebastian Bruch}, \bibinfo{person}{Nadav Golbandi},
  \bibinfo{person}{Michael Bendersky}, {and} \bibinfo{person}{Marc Najork}.}
  \bibinfo{year}{2019}\natexlab{}.
\newblock \showarticletitle{Learning Groupwise Multivariate Scoring Functions
  Using Deep Neural Networks}. In \bibinfo{booktitle}{\emph{Proceedings of the
  2019 ACM SIGIR International Conference on Theory of Information Retrieval}}.
  ACM, \bibinfo{pages}{85--92}.
\newblock


\bibitem[\protect\citeauthoryear{Beel, Genzmehr, Langer, N{\"u}rnberger, and
  Gipp}{Beel et~al\mbox{.}}{2013}]%
        {beel2013comparative}
\bibfield{author}{\bibinfo{person}{Joeran Beel}, \bibinfo{person}{Marcel
  Genzmehr}, \bibinfo{person}{Stefan Langer}, \bibinfo{person}{Andreas
  N{\"u}rnberger}, {and} \bibinfo{person}{Bela Gipp}.}
  \bibinfo{year}{2013}\natexlab{}.
\newblock \showarticletitle{A comparative analysis of offline and online
  evaluations and discussion of research paper recommender system evaluation}.
  In \bibinfo{booktitle}{\emph{Proceedings of the international workshop on
  reproducibility and replication in recommender systems evaluation}}.
  \bibinfo{pages}{7--14}.
\newblock


\bibitem[\protect\citeauthoryear{Burges}{Burges}{2010}]%
        {burges2010ranknet}
\bibfield{author}{\bibinfo{person}{Christopher~JC Burges}.}
  \bibinfo{year}{2010}\natexlab{}.
\newblock \showarticletitle{From ranknet to lambdarank to lambdamart: An
  overview}.
\newblock \bibinfo{journal}{\emph{Learning}} \bibinfo{volume}{11},
  \bibinfo{number}{23-581} (\bibinfo{year}{2010}), \bibinfo{pages}{81}.
\newblock


\bibitem[\protect\citeauthoryear{Burges, Shaked, Renshaw, Lazier, Deeds,
  Hamilton, and Hullender}{Burges et~al\mbox{.}}{2005}]%
        {Burges:ranknet}
\bibfield{author}{\bibinfo{person}{Christopher J.~C. Burges},
  \bibinfo{person}{Tal Shaked}, \bibinfo{person}{Erin Renshaw},
  \bibinfo{person}{Ari Lazier}, \bibinfo{person}{Matt Deeds},
  \bibinfo{person}{Nicole Hamilton}, {and} \bibinfo{person}{Gregory~N.
  Hullender}.} \bibinfo{year}{2005}\natexlab{}.
\newblock \showarticletitle{Learning to rank using gradient descent}. In
  \bibinfo{booktitle}{\emph{Machine Learning, Proceedings of the Twenty-Second
  International Conference {(ICML} 2005)}}. \bibinfo{pages}{89--96}.
\newblock
\urldef\tempurl%
\url{https://doi.org/10.1145/1102351.1102363}
\showDOI{\tempurl}


\bibitem[\protect\citeauthoryear{Cao, Qin, Liu, Tsai, and Li}{Cao
  et~al\mbox{.}}{2007}]%
        {cao07list}
\bibfield{author}{\bibinfo{person}{Zhe Cao}, \bibinfo{person}{Tao Qin},
  \bibinfo{person}{Tie{-}Yan Liu}, \bibinfo{person}{Ming{-}Feng Tsai}, {and}
  \bibinfo{person}{Hang Li}.} \bibinfo{year}{2007}\natexlab{}.
\newblock \showarticletitle{Learning to rank: from pairwise approach to
  listwise approach}. In \bibinfo{booktitle}{\emph{Machine Learning,
  Proceedings of the Twenty-Fourth International Conference {(ICML} 2007),
  Corvallis, Oregon, USA, June 20-24, 2007}}. \bibinfo{pages}{129--136}.
\newblock
\urldef\tempurl%
\url{https://doi.org/10.1145/1273496.1273513}
\showDOI{\tempurl}


\bibitem[\protect\citeauthoryear{Cossock and Zhang}{Cossock and Zhang}{2008}]%
        {cossock08point}
\bibfield{author}{\bibinfo{person}{David Cossock} {and} \bibinfo{person}{Tong
  Zhang}.} \bibinfo{year}{2008}\natexlab{}.
\newblock \showarticletitle{Statistical Analysis of Bayes Optimal Subset
  Ranking}.
\newblock \bibinfo{journal}{\emph{{IEEE} Trans. Information Theory}}
  \bibinfo{volume}{54}, \bibinfo{number}{11} (\bibinfo{year}{2008}),
  \bibinfo{pages}{5140--5154}.
\newblock
\urldef\tempurl%
\url{https://doi.org/10.1109/TIT.2008.929939}
\showDOI{\tempurl}


\bibitem[\protect\citeauthoryear{Dacrema, Cremonesi, and Jannach}{Dacrema
  et~al\mbox{.}}{2019}]%
        {dacrema2019we}
\bibfield{author}{\bibinfo{person}{Maurizio~Ferrari Dacrema},
  \bibinfo{person}{Paolo Cremonesi}, {and} \bibinfo{person}{Dietmar Jannach}.}
  \bibinfo{year}{2019}\natexlab{}.
\newblock \showarticletitle{Are we really making much progress? A worrying
  analysis of recent neural recommendation approaches}. In
  \bibinfo{booktitle}{\emph{Proceedings of the 13th ACM Conference on
  Recommender Systems}}. \bibinfo{pages}{101--109}.
\newblock


\bibitem[\protect\citeauthoryear{Finn, Levine, and Abbeel}{Finn
  et~al\mbox{.}}{2016}]%
        {finn2016guided}
\bibfield{author}{\bibinfo{person}{Chelsea Finn}, \bibinfo{person}{Sergey
  Levine}, {and} \bibinfo{person}{Pieter Abbeel}.}
  \bibinfo{year}{2016}\natexlab{}.
\newblock \showarticletitle{Guided cost learning: Deep inverse optimal control
  via policy optimization}. In \bibinfo{booktitle}{\emph{International
  Conference on Machine Learning}}. \bibinfo{pages}{49--58}.
\newblock


\bibitem[\protect\citeauthoryear{Goodfellow, Pouget{-}Abadie, Mirza, Xu,
  Warde{-}Farley, Ozair, Courville, and Bengio}{Goodfellow
  et~al\mbox{.}}{2014}]%
        {goodfellow15gan}
\bibfield{author}{\bibinfo{person}{Ian~J. Goodfellow}, \bibinfo{person}{Jean
  Pouget{-}Abadie}, \bibinfo{person}{Mehdi Mirza}, \bibinfo{person}{Bing Xu},
  \bibinfo{person}{David Warde{-}Farley}, \bibinfo{person}{Sherjil Ozair},
  \bibinfo{person}{Aaron~C. Courville}, {and} \bibinfo{person}{Yoshua Bengio}.}
  \bibinfo{year}{2014}\natexlab{}.
\newblock \showarticletitle{Generative Adversarial Nets}. In
  \bibinfo{booktitle}{\emph{Annual Conference on Neural Information Processing
  Systems 2014, December 8-13 2014, Montreal}}. \bibinfo{pages}{2672--2680}.
\newblock


\bibitem[\protect\citeauthoryear{Ho and Ermon}{Ho and Ermon}{2016}]%
        {ho16gail}
\bibfield{author}{\bibinfo{person}{Jonathan Ho} {and} \bibinfo{person}{Stefano
  Ermon}.} \bibinfo{year}{2016}\natexlab{}.
\newblock \showarticletitle{Generative Adversarial Imitation Learning}. In
  \bibinfo{booktitle}{\emph{Annual Conference on Neural Information Processing
  Systems 2016, December 5-10, 2016, Barcelona, Spain}}.
  \bibinfo{pages}{4565--4573}.
\newblock


\bibitem[\protect\citeauthoryear{Hu, Koren, and Volinsky}{Hu
  et~al\mbox{.}}{2008}]%
        {hu2008collaborative}
\bibfield{author}{\bibinfo{person}{Yifan Hu}, \bibinfo{person}{Yehuda Koren},
  {and} \bibinfo{person}{Chris Volinsky}.} \bibinfo{year}{2008}\natexlab{}.
\newblock \showarticletitle{Collaborative filtering for implicit feedback
  datasets}. In \bibinfo{booktitle}{\emph{2008 Eighth IEEE International
  Conference on Data Mining}}. Ieee, \bibinfo{pages}{263--272}.
\newblock


\bibitem[\protect\citeauthoryear{Huzhang, Pang, Gao, Liu, Shen, Zhou, Da, Zeng,
  Yu, Yu, et~al\mbox{.}}{Huzhang et~al\mbox{.}}{2020}]%
        {huzhang2020aliexpress}
\bibfield{author}{\bibinfo{person}{Guangda Huzhang}, \bibinfo{person}{Zhen-Jia
  Pang}, \bibinfo{person}{Yongqing Gao}, \bibinfo{person}{Yawen Liu},
  \bibinfo{person}{Weijie Shen}, \bibinfo{person}{Wen-Ji Zhou},
  \bibinfo{person}{Qing Da}, \bibinfo{person}{An-Xiang Zeng},
  \bibinfo{person}{Han Yu}, \bibinfo{person}{Yang Yu}, {et~al\mbox{.}}}
  \bibinfo{year}{2020}\natexlab{}.
\newblock \showarticletitle{AliExpress Learning-To-Rank: Maximizing Online
  Model Performance without Going Online}.
\newblock \bibinfo{journal}{\emph{arXiv preprint arXiv:2003.11941}}
  (\bibinfo{year}{2020}).
\newblock


\bibitem[\protect\citeauthoryear{Ie, Hsu, Mladenov, Jain, Narvekar, Wang, Wu,
  and Boutilier}{Ie et~al\mbox{.}}{2019}]%
        {ie2019recsim}
\bibfield{author}{\bibinfo{person}{Eugene Ie}, \bibinfo{person}{Chih-wei Hsu},
  \bibinfo{person}{Martin Mladenov}, \bibinfo{person}{Vihan Jain},
  \bibinfo{person}{Sanmit Narvekar}, \bibinfo{person}{Jing Wang},
  \bibinfo{person}{Rui Wu}, {and} \bibinfo{person}{Craig Boutilier}.}
  \bibinfo{year}{2019}\natexlab{}.
\newblock \showarticletitle{RecSim: A Configurable Simulation Platform for
  Recommender Systems}.
\newblock \bibinfo{journal}{\emph{arXiv preprint arXiv:1909.04847}}
  (\bibinfo{year}{2019}).
\newblock


\bibitem[\protect\citeauthoryear{Joachims}{Joachims}{2002}]%
        {joachims02pair}
\bibfield{author}{\bibinfo{person}{Thorsten Joachims}.}
  \bibinfo{year}{2002}\natexlab{}.
\newblock \showarticletitle{Optimizing search engines using clickthrough data}.
  In \bibinfo{booktitle}{\emph{Proceedings of the Eighth {ACM} {SIGKDD}
  International Conference on Knowledge Discovery and Data Mining, July 23-26,
  2002, Edmonton, Alberta, Canada}}. \bibinfo{pages}{133--142}.
\newblock
\urldef\tempurl%
\url{https://doi.org/10.1145/775047.775067}
\showDOI{\tempurl}


\bibitem[\protect\citeauthoryear{Joachims, Swaminathan, and Schnabel}{Joachims
  et~al\mbox{.}}{2017}]%
        {joachims2017unbiased}
\bibfield{author}{\bibinfo{person}{Thorsten Joachims}, \bibinfo{person}{Adith
  Swaminathan}, {and} \bibinfo{person}{Tobias Schnabel}.}
  \bibinfo{year}{2017}\natexlab{}.
\newblock \showarticletitle{Unbiased learning-to-rank with biased feedback}. In
  \bibinfo{booktitle}{\emph{Proceedings of the Tenth ACM International
  Conference on Web Search and Data Mining}}. \bibinfo{pages}{781--789}.
\newblock


\bibitem[\protect\citeauthoryear{Li, Burges, and Wu}{Li et~al\mbox{.}}{2007}]%
        {li07point}
\bibfield{author}{\bibinfo{person}{Ping Li}, \bibinfo{person}{Christopher J.~C.
  Burges}, {and} \bibinfo{person}{Qiang Wu}.} \bibinfo{year}{2007}\natexlab{}.
\newblock \showarticletitle{McRank: Learning to Rank Using Multiple
  Classification and Gradient Boosting}. In
  \bibinfo{booktitle}{\emph{Proceedings of the Twenty-First Annual Conference
  on Neural Information Processing Systems, Vancouver, British Columbia,
  Canada, December 3-6, 2007}}. \bibinfo{pages}{897--904}.
\newblock


\bibitem[\protect\citeauthoryear{Mao, Li, Li, Chen, Wang, and Deng}{Mao
  et~al\mbox{.}}{2020}]%
        {mao2020www}
\bibfield{author}{\bibinfo{person}{Huiqiang Mao}, \bibinfo{person}{Yanzhi Li},
  \bibinfo{person}{Chenliang Li}, \bibinfo{person}{Di Chen},
  \bibinfo{person}{Xiaoqing Wang}, {and} \bibinfo{person}{Yuming Deng}.}
  \bibinfo{year}{2020}\natexlab{}.
\newblock \showarticletitle{PARS: Peers-Aware Recommender System}. In
  \bibinfo{booktitle}{\emph{Proceedings of The Web Conference 2020}} (Taipei,
  Taiwan) \emph{(\bibinfo{series}{WWW '20})}. \bibinfo{publisher}{Association
  for Computing Machinery}, \bibinfo{address}{New York, NY, USA},
  \bibinfo{pages}{2606–2612}.
\newblock
\showISBNx{9781450370233}
\urldef\tempurl%
\url{https://doi.org/10.1145/3366423.3380013}
\showDOI{\tempurl}


\bibitem[\protect\citeauthoryear{McNee, Riedl, and Konstan}{McNee
  et~al\mbox{.}}{2006}]%
        {mcnee2006being}
\bibfield{author}{\bibinfo{person}{Sean~M McNee}, \bibinfo{person}{John Riedl},
  {and} \bibinfo{person}{Joseph~A Konstan}.} \bibinfo{year}{2006}\natexlab{}.
\newblock \showarticletitle{Being accurate is not enough: how accuracy metrics
  have hurt recommender systems}. In \bibinfo{booktitle}{\emph{CHI'06 extended
  abstracts on Human factors in computing systems}}.
  \bibinfo{pages}{1097--1101}.
\newblock


\bibitem[\protect\citeauthoryear{Rendle, Freudenthaler, Gantner, and
  Schmidt-Thieme}{Rendle et~al\mbox{.}}{2012}]%
        {rendle2012bpr}
\bibfield{author}{\bibinfo{person}{Steffen Rendle}, \bibinfo{person}{Christoph
  Freudenthaler}, \bibinfo{person}{Zeno Gantner}, {and} \bibinfo{person}{Lars
  Schmidt-Thieme}.} \bibinfo{year}{2012}\natexlab{}.
\newblock \showarticletitle{BPR: Bayesian personalized ranking from implicit
  feedback}.
\newblock \bibinfo{journal}{\emph{arXiv preprint arXiv:1205.2618}}
  (\bibinfo{year}{2012}).
\newblock


\bibitem[\protect\citeauthoryear{Rohde, Bonner, Dunlop, Vasile, and
  Karatzoglou}{Rohde et~al\mbox{.}}{2018}]%
        {rohde2018recogym}
\bibfield{author}{\bibinfo{person}{David Rohde}, \bibinfo{person}{Stephen
  Bonner}, \bibinfo{person}{Travis Dunlop}, \bibinfo{person}{Flavian Vasile},
  {and} \bibinfo{person}{Alexandros Karatzoglou}.}
  \bibinfo{year}{2018}\natexlab{}.
\newblock \showarticletitle{Recogym: A reinforcement learning environment for
  the problem of product recommendation in online advertising}.
\newblock \bibinfo{journal}{\emph{arXiv preprint arXiv:1808.00720}}
  (\bibinfo{year}{2018}).
\newblock


\bibitem[\protect\citeauthoryear{Rossetti, Stella, and Zanker}{Rossetti
  et~al\mbox{.}}{2016}]%
        {rossetti2016contrasting}
\bibfield{author}{\bibinfo{person}{Marco Rossetti}, \bibinfo{person}{Fabio
  Stella}, {and} \bibinfo{person}{Markus Zanker}.}
  \bibinfo{year}{2016}\natexlab{}.
\newblock \showarticletitle{Contrasting offline and online results when
  evaluating recommendation algorithms}. In
  \bibinfo{booktitle}{\emph{Proceedings of the 10th ACM conference on
  recommender systems}}. \bibinfo{pages}{31--34}.
\newblock


\bibitem[\protect\citeauthoryear{Shi, Yu, Da, Chen, and Zeng}{Shi
  et~al\mbox{.}}{2019}]%
        {shi2019virtual}
\bibfield{author}{\bibinfo{person}{Jing-Cheng Shi}, \bibinfo{person}{Yang Yu},
  \bibinfo{person}{Qing Da}, \bibinfo{person}{Shi-Yong Chen}, {and}
  \bibinfo{person}{An-Xiang Zeng}.} \bibinfo{year}{2019}\natexlab{}.
\newblock \showarticletitle{Virtual-taobao: Virtualizing real-world online
  retail environment for reinforcement learning}. In
  \bibinfo{booktitle}{\emph{Proceedings of the AAAI Conference on Artificial
  Intelligence}}, Vol.~\bibinfo{volume}{33}. \bibinfo{pages}{4902--4909}.
\newblock


\bibitem[\protect\citeauthoryear{Xia, Liu, Wang, Zhang, and Li}{Xia
  et~al\mbox{.}}{2008}]%
        {xia08list}
\bibfield{author}{\bibinfo{person}{Fen Xia}, \bibinfo{person}{Tie{-}Yan Liu},
  \bibinfo{person}{Jue Wang}, \bibinfo{person}{Wensheng Zhang}, {and}
  \bibinfo{person}{Hang Li}.} \bibinfo{year}{2008}\natexlab{}.
\newblock \showarticletitle{Listwise approach to learning to rank: theory and
  algorithm}. In \bibinfo{booktitle}{\emph{Machine Learning, Proceedings of the
  Twenty-Fifth International Conference {(ICML} 2008), Helsinki, Finland, June
  5-9, 2008}}. \bibinfo{pages}{1192--1199}.
\newblock
\urldef\tempurl%
\url{https://doi.org/10.1145/1390156.1390306}
\showDOI{\tempurl}


\bibitem[\protect\citeauthoryear{Yu, Liu, Ye, Cheng, Chen, and Ma}{Yu
  et~al\mbox{.}}{2020}]%
        {yu2020collaborative}
\bibfield{author}{\bibinfo{person}{Runlong Yu}, \bibinfo{person}{Qi Liu},
  \bibinfo{person}{Yuyang Ye}, \bibinfo{person}{Mingyue Cheng},
  \bibinfo{person}{Enhong Chen}, {and} \bibinfo{person}{Jianhui Ma}.}
  \bibinfo{year}{2020}\natexlab{}.
\newblock \showarticletitle{Collaborative List-and-Pairwise Filtering from
  Implicit Feedback}.
\newblock \bibinfo{journal}{\emph{IEEE Transactions on Knowledge and Data
  Engineering}} (\bibinfo{year}{2020}).
\newblock


\end{thebibliography}

%\appendix









\end{document}